# GMOTE: Gaussian based minority oversampling technique for imbalanced classification adapting tail probability of outliers

Seung Jee Yang[a], Kyung Joon Cha[b*]

[a]*Department of applied statistics and Institute for Convergence of Basic Sciences, Hanyang University, Seoul, 04763, Republic of Korea*

[b]*Department of mathematics and Institute for Convergence of Basic Sciences, Hanyang University, Seoul, 04763, Republic of Korea*

**Abstract**

Classification of imbalanced data is one of the common problems in the recent field of data mining. Imbalanced data substantially affects the performance of standard classification models. Data-level approaches mainly use the oversampling methods to solve the problem, such as synthetic minority oversampling Technique (SMOTE). However, since the methods such as SMOTE generate instances by linear interpolation, synthetic data space may look like a polygonal. Also, the oversampling methods generate outliers of the minority class. In this paper, we proposed Gaussian based minority oversampling technique (GMOTE) with a statistical perspective for imbalanced datasets. To avoid linear interpolation and to consider outliers, this proposed method generates instances by the Gaussian Mixture Model. Motivated by clustering-based multivariate Gaussian outlier score (CMGOS), we propose to adapt tail probability of instances through the Mahalanobis distance to consider local outliers. The experiment was carried out on a representative set of benchmark datasets. The performance of the GMOTE is compared with other methods such as SMOTE. When the GMOTE is combined with classification and regression tree (CART) or support vector machine (SVM), it shows better accuracy and F1-Score. Experimental results demonstrate the robust performance.

*Keywords:* Imbalanced data, Outliers, Gaussian mixture model, Tail probability

## 1. Introduction

With the recent development of information technology entering the age of big data, it is becoming common to process huge amounts of data. To deal with huge amounts of data, machine learning, pattern recognition, and data mining methodologies have been used widely in many fields. However, there are problems in various domains such as credit card fraud [6], software defect prediction [43], text mining [33], pattern recognition [46], and gene mining [48] if data follows a biased distribution. It is important to correctly classify the instances of the minority class on the domains mentioned because the cost of misclassification for the minority class is relatively greater than for the majority class [15,51]. Also, since many standard classifiers such as logistic regression, decision tree, and support vector machine (SVM) aim at maximizing classification accuracy, the minority class is highly likely to be classified as a majority class [42].

[*]Corresponding author.

*Email addresses:* kjcha@hanyang.ac.kr

As a method to solve these problems, data-level approach, cost-sensitive learning, and ensemble technique are proposed [34]. Data-level approaches include resampling, feature selection, and extraction. Cost-sensitive learning allocates a larger penalty for misclassifying minority class instances with respect to majority class instances. Finally, the ensemble techniques are the methods of improving performance by aggregating several classifiers with low performance.

Generally, the data-level approaches have been studied [22]. As resampling methods, there are oversampling, undersampling, and hybrid-sampling. Oversampling replicates instances from the minority class to fit the size of majority class data. Undersampling randomly removes some instances of the majority class to fit the size of the minority class data. The hybrid method considers over and undersampling at the same time. Since the hybrid method balances the number of instances between the minority class and the majority class, it achieves a significant improvement in classification performance [2,11].

Since these methods are simple preprocessing techniques, there are some drawbacks. Oversampling may lead to overfitting because it replicates instances of the minority class. On the other hand, the disadvantage of undersampling is a loss of potentially important information for the majority class. When the number of minority instances is small, undersampling may limit the classification performance [15].

One straightforward solution to these problems is the synthetic minority oversampling technique (SMOTE) [12] that generates synthetic instances by linearly interpolating $k$-nearest neighbors. However, if we do not consider the outlier of the minority class instances, synthetic outliers can occur. The outliers or generated outliers may reduce the classification performance for both the minority and majority classes [31].

There are several variations of SMOTE that aim to handle the weaknesses of the original algorithm, such as Borderline-SMOTE [23], Safe-level SMOTE [9], DBSMOTE [10], and Cluster-SMOTE [13]. However, since they generate synthetic instances by linear interpolation, there is a drawback that the synthetic data space is the tied region [50]. Due to this space limitation, it could overfit the given data. Since the distance-based approaches do not consider the statistical characteristics of the minority class, it is difficult to identify the outlier depending on the definition of distance or the parameter setting of the sampling process.

Consequently, some methods such as probability density function estimation based oversampling (PDFOS) [17] and Gaussian prior based adaptive synthetic sampling (GA-SMOTE) [50] have been proposed recently. They endeavor to avoid generating synthetic instances by following data distribution instead of linear interpolation. Despite the efforts of PDFOS and GA-SMOTE, we may face overfitting. Because synthetic instances are generated locally near the sample, but samples may exist globally. Recently, the Gaussian oversampling which generates synthetic instances on the rectangle sampling region obtained from dimension reduction was proposed [36]. However, this approach may be inappropriate from a statistical perspective if there are correlations between variables. When the domain of datasets could have correlations, we should consider the contour of the elliptical shape rather than the rectangular shape.

Therefore, we propose a Gaussian based minority oversampling technique (GMOTE) to address the binary imbalanced classification problems. In detail, the main contributions of this work are as follows.

(1) We use Gaussian mixture model (GMM) to approximate the distribution of the minority class or subgroup of minority class as a statistical approach. To this end, GMM is obtained through EM algorithm. Also, the number of GMM's components is automatically decided through the Bayesian Information Criterion (BIC) [16,41].

(2) We improve the robustness of classifier performance using tail probability to handle the outlying instances. On sample space whose dimension is greater than two, we use the Mahalanobis distance between the center of GMM's components and the minority instance. Since the Mahalanobis distance is a well-known distance available through the means and standard deviations of the multivariate distribution, it is used to detect outliers by calculating the tail probability indirectly in the distribution.

(3) The proposed GMOTE generates the artificial instances using the random number generator to ensure more natural in the generation process. Thus, the generated artificial samples follow the distribution of minority and do not lie in linearly tied artificial regions.

## 2. Preliminaries

*2.1. Review of oversampling methods for imbalanced dataset*

We briefly introduce widely used oversampling methods to show the differences in the way of instance generation with GMOTE. The SMOTE method widely referred among the oversampling methods for class imbalance problems. The SMOTE generates synthetic instance $x_{syn}$ using linear interpolation between minority instance $x_i$ and its *k*-nearest neighbor $\hat{x}_i$. A new synthetic instance can be described as

$$x_{syn} = x_i + u \times (\hat{x}_i - x_i), \qquad u \sim Unif(0,1),$$

where $u$ is a random variable of the uniform distribution. However, SMOTE causes over-fitting because SMOTE places the synthetic instance on the same line with high probability. Thus, the SMOTE increases a recall but decreases a specificity. Also, SMOTE generates the synthetic outlier when the outlier of minority class exists.

The Borderline-SMOTE operates only on borderline instances in the overlapping region. This method defines a minority instance whose $k$-nearest neighbors are all majority class as "Noise", and a minority instance whose number of minority neighbors is high as "Safe". And "Danger" is neither "Noise" nor "Safe," i.e., "Danger" is close to the decision boundary and is in danger of misclassification. Borderline-SMOTE uses only "Danger" instances. Thus, Borderline-SMOTE is not vulnerable to noise. However, the rate of accuracy for the majority class (i.e., specificity) is disappointing since the classifier misclassifies instances as minority class on borderline.

The Safe-level SMOTE which is motivated by Borderline-SMOTE calculates the safe-level of the minority instance. The safe level of an instance is the number of minority instances in its $k$-nearest neighbors. And the safe level ratio is the ratio of the safe level of the minority instance to the safe level of its $k$-nearest neighbor. Then the synthetic instances are close to the minority instance with the higher safe level ratio. However, this method causes a relatively lower recall since synthetic instances are sparse in an overlapping region.

Cluster-SMOTE uses $k$-means to divide minority class into multiple clusters and then apply SMOTE in each cluster. The goal of this method is to boost class regions by generating instances within naturally occurring clusters of the minority class [15]. As a clustering method, $k$-means is simple, flexible, and suitable in a large dataset. But the drawbacks of $k$-means come from the assumption that each cluster is a spherical shape with a similar number of instances. Thus, Cluster-SMOTE is also vulnerable to the outlier.

The DBSMOTE uses a density-based clustering algorithm called DBSCAN. So, the outlying instances are easy to handle by DBSCAN. This method makes the graph that constructs the shortest path from the pseudo-centroid to minority instances for each cluster. Then the synthetic instances are generated on the edge of graphs. The result of the process is that synthetic instances are dense around the core of each cluster. But the shape of synthetic instances is close to the star graph. This means that there is a high possibility of overfitting.

Unlike nearest neighborhood methods, Radial-based oversampling [32] uses radial-based functions to oversample from the potential surface. The potential surface represents the degree to which it belongs to the minority or majority class and is calculated by estimating the local distribution of the two classes. The minority instances are generated in the area where the potential of the minority class is higher than the majority class. It is difficult to generate instances in overlapping areas since the potentials of the two classes are similar. However, although the majority's influence is small, minority instances may occur in areas far from the minority class.

In summary, there has been a recent research approach to the improvement of imbalanced dataset oversampling. Nevertheless, overfitting occurs in classification since oversampling methods generate synthetic instances using linear interpolation or around existing minority instances. So it is hard to handle outlying instance for some oversampling methods.

*2.2. Detecting the Outliers*

Since the proposed method uses the tail probability of outlier to generate instances, we need to see the definitions of the outlier in detail. Probably the first definition of the outlier is attributable to Grubbs [19], "An outlying observation, or outlier, is one that appears to deviate markedly from other members of the sample in which it occurs." In the data mining and statistics literature, outliers are also referred to as anomalies, abnormalities, or deviants [1]. They are often used interchangeably in the context of outlier detection though there are some differences between outliers and anomalies [45]. While outliers tend to lay emphasis on statistical rarity and deviation, anomalies are caused by a different underlying generative mechanism [7]. In this paper, we use outliers as a parent concept that contains anomalies.

Recently, number of outlier detection approaches have been proposed. Based on the techniques used, these approaches are classified as follows: distribution-based, distance-based, density-based, and clustering-based methods [49]. Distribution-based methods [5,24,40] depend on the statistical models, i.e., it assumes instances that have low probability to belong to the statistical models are identified as outliers. However, distribution-based methods have a key drawback. It is difficult to use in practical applications if there is no prior knowledge of the data distribution [49]. Distance-based methods [4,29,30,39] define an instance as an outlier with respect to parameters $k$ and $d$ if there are less than $k$ points within the distance $d$ from the point $x$. This definition is suitable if there is no prior knowledge of the data distribution. However, these approaches require the value of parameters $k$ and $d$. Density-based methods [8,27,37] identify the instances as local outliers by computing the density of the local neighborhood called Local Outlier Factor (LOF). However, the LOF for each instance is expensive in that the distance between the instances must be calculated.

Finally, there are two types of clustering-based methods used to construct an outlier score, i.e. the distance to the center of the cluster and the size of clusters [7]. The former assumes that the outliers are far from the centers while the normal data instances are close to them [20,28]. In particular, the proposed method is motivated by the clustering-based multivariate Gaussian outlier score (CMGOS) [18]. In the CMGOS, the local density estimation is attained through the estimate of the multivariate Gaussian model, and the Mahalanobis distance performs as the basis for computing outlier scores. As the first step, $k$-means clustering is applied to separate the dataset into large and small clusters. Then the covariance matrix is computed for each cluster. As the final step, the CMGOS score is computed as a fraction, which is the Mahalanobis distance of an instance to its nearest cluster center divided by the chi-square distribution with a certain confidence interval. The latter considers that the cluster of outliers is sparse and small unlike the cluster of normal data instances [26,47]. Even though clustering-based methods are unsupervised methods, their main purpose is to find clusters which avoids adverse effect on the final clustering result. Especially, drawbacks of the CMGOS are that $k$-means clustering needs the number of clusters, $k$, and does not reflect covariance or correlation of variables.

## 3. Proposed method

In this section, we present the algorithm of GMOTE. GMOTE uses a GMM clustering algorithm in conjunction with oversampling via random number generator to address the imbalanced dataset. Its focus is to reflect the statistical distribution of datasets and to consider local outliers. In addition, it manages to avoid the generation of the artificial outlier and synthetic instances by linear interpolation. GMOTE consists of three functions: clustering, detecting outliers and oversampling. The clustering with GMM is performed via the EM algorithm. GMOTE detects whether the instance is outlier or not on each cluster by the Mahalanobis distance. After detectin outliers, GMM is performed again without outliers. Then, the new GMM generates artificial instances. The framework of the proposed GMOTE algorithm is shown in Fig. 1.

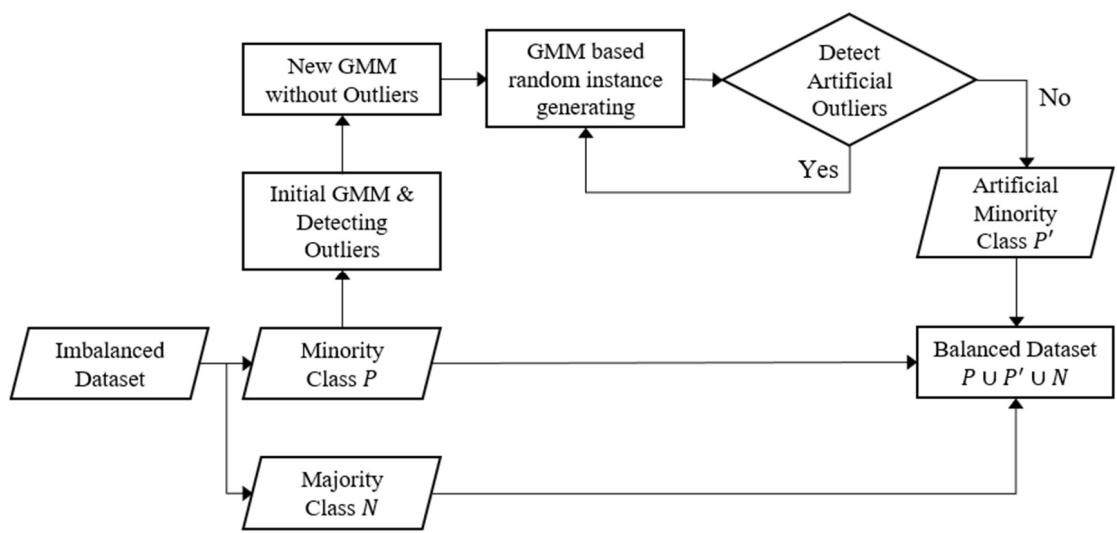

Fig. 1. The flowchart of GMOTE.

*3.1 Estimation of GMM via EM algorithm*

It is important to construct a reasonable probability distribution model using the statistical characteristics of the minority class [36]. In statistics and many fields of applied science, the Gaussian distribution has been classically used to describe data distributions.

Given a training dataset $D = \{(x_i, y_i)\}, i = 1, 2, \dots, N$, where $x_i = (x_{i1}, x_{i2}, \dots, x_{iM})^T$ is the $i$th instance with $M$-dimensional feature. GMM is a weighted sum of $C$ component Gaussian distributions as below,

$$p(x|\mu, \Sigma, \pi) = \sum_{c=1}^{C} \pi_c \mathcal{N}(x|\mu_c, \Sigma_c),$$

and each component's density is $M$-variate Gaussian function of the form,

$$\mathcal{N}(x|\mu_c, \Sigma_c) = \frac{1}{(2\pi)^{\frac{M}{2}} |\Sigma_c|^{\frac{1}{2}}} \exp\left[-\frac{1}{2}(x-\mu_c)^T \Sigma_c^{-1}(x-\mu_c)\right]$$

with mean vector $\mu_c$, and covariance matrix $\Sigma_c$. The component weight $\pi_c$ is the degree of importance that denotes the contribution of the $c$th component to the overall distributive characteristics of dataset, such that $\sum_{c=1}^{C} \pi_c = 1$.

To estimate the distributions of minority class by the GMM, we consider the proper parameter such as the number of components or the mean and covariance of each component. Since that parameter is an unknown or latent variable, the instances of minority class can be viewed as incomplete data. To compute directly the maximum likelihood estimator (MLE) is complicated, so MLE of a finite mixture model is frequently solved via the EM algorithm [14] or its variants. The EM algorithm seeks to find the MLE of a finite mixture model by iteratively applying as follows: estimation the conditional expectation of the complete data log-likelihood given the observed data and the current parameter estimates (E-step) and re-estimation the parameters using the current posterior probabilities (M-step).

Meanwhile, there is still a problem that the number of GMM's components is not determined. There are several methods to determine the proper number of components [16]. We use the information criteria that are based on a penalized form of the log-likelihood since we need to maintain consistency concerning the view of probability. The BIC [41, 16] is widely used in the context of GMMs and takes the form as

$$\text{BIC} = -2 \log L(\widehat{\Psi}|x) + \nu \log C,$$

where $L(\widehat{\Psi}|x)$ is the likelihood at the MLE $\widehat{\Psi}$ formed from the observed data $x$ and the number of estimated parameters $\nu$. We select the number of components with the lowest BIC.

*3.2. Calculation of tail probability to consider the local outliers*

The purpose of this study is to propose a GMOTE for statistical oversampling. GMOTE can approach the distribution of minority class instances while considering the tail probability of outliers of the minority class. Unlike the other methods that consider the outlier through frequency, GMOTE calculates the local tail probability of minority class outlying instance by the Mahalanobis distance proposed by P. C. Mahalanobis [35]. In the univariate outlier detection, we can calculate the tail probability easily. But it is difficult to calculate the tail probability in the multivariate outlier detection. For calculating tail probability on sample space whose dimension is greater than two, we use the Mahalanobis distance to consider the correlation of variables.

The Mahalanobis distance is the distance between an instance $x$ and the multivariate distribution. It is used to measure how distant an instance $x$ is from $\mu$ and $\Sigma$. Since the Mahalanobis distance considers the covariance between features, it is used on multivariate data. The Mahalanobis distance is given by below,

$$d(x, \mu, \Sigma) = \sqrt{(x-\mu)^T \Sigma^{-1} (x-\mu)}.$$

The test statistic for the Mahalanobis distance is defined as Hotelling's $T$-squared statistic ($T^2$) which is related to the $F$-distribution proposed by H. Hotelling [25] and it is given as

$$T_i^2 = d(x_i, \bar{x}, \hat{\Sigma})^2 \sim \frac{M(N-1)}{N-M} F_{M,N-M}.$$

Here, $\bar{x}$ is the sample mean vector, $\hat{\Sigma}$ is the sample covariance matrix, and $F_{M,N-M}$ is the $F$-distribution with parameters $M$ and $N-M$. For large samples, $d(x_i, \mu, \Sigma)^2$ follows a Chi-squared distribution with $M$ degrees of freedom.

The Mahalanobis distance is used to detect outliers [18,21,38]. To identify local outlier, we calculate the $p$-value of instances via the Mahalanobis distance $d(x_i, \mu, \Sigma)^2$ for each GMM's component as below:

$$p_c^{(i)} = \mathrm{P}(d^2 > d(x_i, \mu_c, \Sigma_c)^2), \qquad p^{(i)} = \min\{p_1^{(i)}, \dots, p_C^{(i)}\},$$

where $p^{(i)}$ is the minimum $p$-value from each component of GMM. If $p^{(i)}$ is less than the cut-off value α, then we deal with $x_i$ as the outlier (Fig. 2).

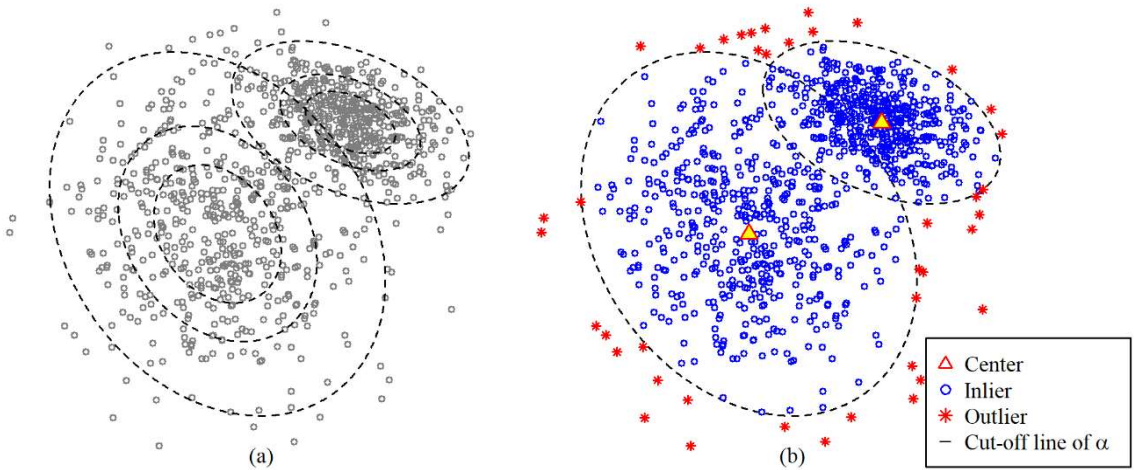

Fig. 2. An example of detecting local outlier. (a) Contour lines indicate each cut-off value for each Gaussian component. (b) Red instances outside of all contours become local outliers.

*3.3. GMM based oversampling minor class considering outlier*

As mentioned before, GMOTE operates in three stages repeatedly. Specifically, the first stage is to achieve clustering by estimating Gaussian Mixture via the EM algorithm. The number of clusters is decided automatically based on BIC. The second stage is to determine the local outliers by calculating local tail probability via the Mahalanobis distance from each component of GMM derived in the previous stage. By comparing local tail probabilities with cut-off value, instances with tail probabilities smaller than cut-off value are judged as the local outlier. Finally, in the oversampling stage, GMOTE generates random artificial instances from each component of GMM.

In order to reduce the influence of outliers on clustering and random number generation, the second stage always operates together in the first and third stages. As shown in Fig. 1, GMM is acquired twice. Unlike the initial GMM, the new GMM is derived from the rest of the instances, except those determined as local outliers, through tail probabilities calculated using the Mahalanobis distance from the initial GMM. Then, artificial instances are generated from each component of the new GMM. In this stage, there is a possibility that artificial instances to be identified as outliers are also generated. Therefore, to avoid generating artificial outliers, measures should be taken to exploit the truncated distribution or to regenerate the generated outliers into inliers. In this paper, we applied the algorithm that regenerates the generated outliers into inliers until there are no artificial outliers to GMOTE. The pseudocode of the described algorithm is presented in Algorithm.

| **Algorithm**: GMOTE |
|---|
| **Input**: |
|     $\alpha$: the cut-off value |
|     $\gamma$: sampling ratio |
|     $P$: the set of minority class |
| **Output**: |
|     $P'$: the artificial minority class instances |
| 1  Train initial GMM via EM from $P$ |
| 2  Calculate minimum tail probability $p_c$ of instances for each initial GMM's component |
| 3  Train new GMM via EM from $P_{del} = \{x | p_c(x) < 1 - \alpha\}$ |
| 4  $P' \leftarrow \emptyset$ |
| 5  **while** $|P'| < \gamma |P|$ **do** |
| 6    $E \leftarrow$ Randomly generated instances from new GMM |
| 7    Calculate minimum tail probability $p_c$ of instances from $E$ for each new GMM's component |
| 8    $P' \leftarrow P' \cup \{x | p_c(x) < 1 - \alpha, \forall x \in E\}$ |
| 9  **end while** |
| 10 **return** $P'$ |

**4. Experimental study**

In this section, comprehensive experiments have been conducted to evaluate the performance empirically. As the first experiment, we compared the GMOTE algorithm with several methods of dealing with imbalanced data under the standard conditions. Then we evaluated the performance of the GMOTE and other methods on typical imbalanced datasets from the Knowledge Extraction based on Evolutionary Learning (KEEL) data repository [3] by using 5-fold cross-validation. Table 1 summarizes the details of these datasets.

**Table 1**
Imbalanced datasets from KEEL.

| ID | Dataset | IR | Samples | Attributes | Train | Test |
| --- | --- | --- | --- | --- | --- | --- |
| KEEL 1 | ecoli0vs1 | 1.86 | 220 | 7 | 176 | 44 |
| KEEL 2 | glass0123vs456 | 3.19 | 214 | 9 | 171 | 43 |
| KEEL 3 | hamberman | 2.68 | 306 | 3 | 244 | 62 |
| KEEL 4 | new thyroid1 | 5.14 | 215 | 5 | 172 | 43 |
| KEEL 5 | pima | 1.9 | 768 | 8 | 614 | 154 |
| KEEL 6 | segement0 | 6.01 | 2308 | 19 | 1846 | 462 |
| KEEL 7 | wisconsin | 1.86 | 683 | 9 | 546 | 137 |
| KEEL 8 | yeast1 | 2.46 | 1484 | 8 | 1187 | 297 |

*4.1. Set-up*

*Methods and parameters*. To verify the effectiveness of the GMOTE algorithm, it was compared with the original data, random oversampling (ROS), SMOTE, Borderline-SMOTE (BLSMOTE), Safe-level SMOTE (SLSMOTE), DBSMOTE, Cluster SMOTE (C-SMOTE), and Radial based oversampling (RBO). Except ROS and RBO, other plain SMOTE methods need to normalize all the attributes to a (0, 1) range based on the original range of the training data. The experiments were implemented using R 3.6.2 and package 'smotefamily' version 1.3.1 [44]. Table 2 summarizes the details of parameter set-up.

**Table 2**
Parameter Set-up

| Methods | Parameter |
| --- | --- |
| Common | sample ratio: 1 |
| SMOTE | K: 3 |
| Borderline-SMOTE | K: 3, C: 5 |
| Safe-level SMOTE | K: 5, C: 5 |
| DBSMOTE | MinPts: 4, eps: Q3 of $5^{th}$ nearest distances |
| C-SMOTE | Cluster: $k$-Means, the number of clusters: 3 |
| RBO | gamma: 1, iteration: 50, step size: 0.05, p=0.05 |
| GMOTE | alpha: 0.05 |

*Performance metrics*. For the performance evaluation metrics of classification, the following metrics are used when the dataset has imbalanced class distribution, but some are not appropriate. We considered the mainly used metrics, namely, accuracy, recall, precision, F1-score, G-mean, and AUC, which can be described by four outcomes: true positive (TP), false positive (FP), false negative (FN), and true negative (TN).

*Classifiers.* To evaluate the effectiveness of the oversampling methods, several different classification algorithms were used as a base classifier, namely: logistic regression, classification and regression tree (CART), support vector machine (SVM). The implementations of the classification algorithms provided in the R packages 'nnet', 'rpart', and 'e1071' were used, together with their default parameters.

*4.2. Preliminary experiment and Experimental result*

Using two-dimensional toy datasets, it is possible to describe how GMOTE may improve the results obtained by plain SMOTE and other SMOTE-like algorithms. We use two examples to show the results of the comparison between methods (Figs. 3 and 4). The gray plus, red cross, and blue dot represent majority class, minority class, and the artificial instance, respectively.

We observe that the generated instances look like polygonal shapes in toy datasets except for RBO and GMOTE. This is because those SMOTE family methods generate instances by linear interpolation. BLSMOTE does not generate artificial instances if there is significantly a lot or very few majority class at the edge of the minority class. In the sample area of minority class, RBO and GMOTE appear to follow the distribution even at a glance. However, RBO generates outliers in the area where the majority class instances are relatively few. We see that GMOTE generates instance inside the minority class like SLSMOTE or DBSMOTE, but also strengthens the border like BLSMOTE. This is because GMOTE does not use the data directly like other methods, but rather generates instances from the distribution through statistic such as sample-mean and sample-covariance.

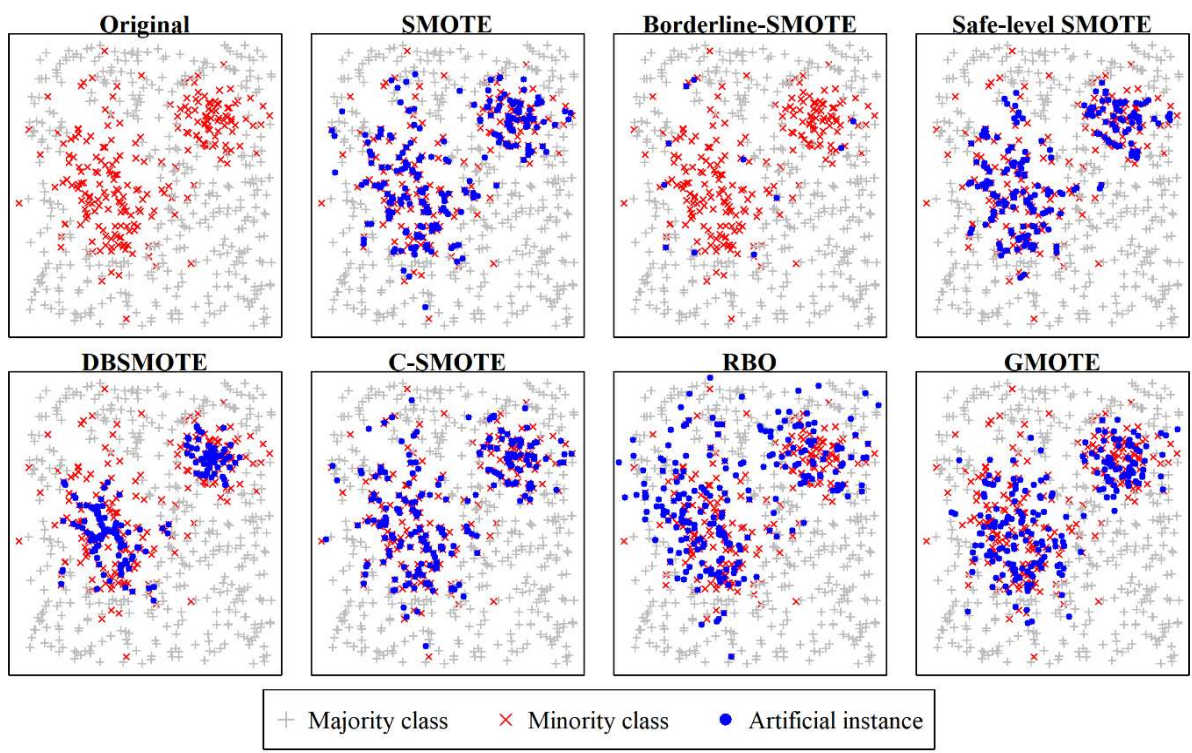

Fig. 3. An example 1 of artificial instances generation by different methods. The majority class instances surround the minority class. The minority class has two clusters that follow the Gaussian distribution.

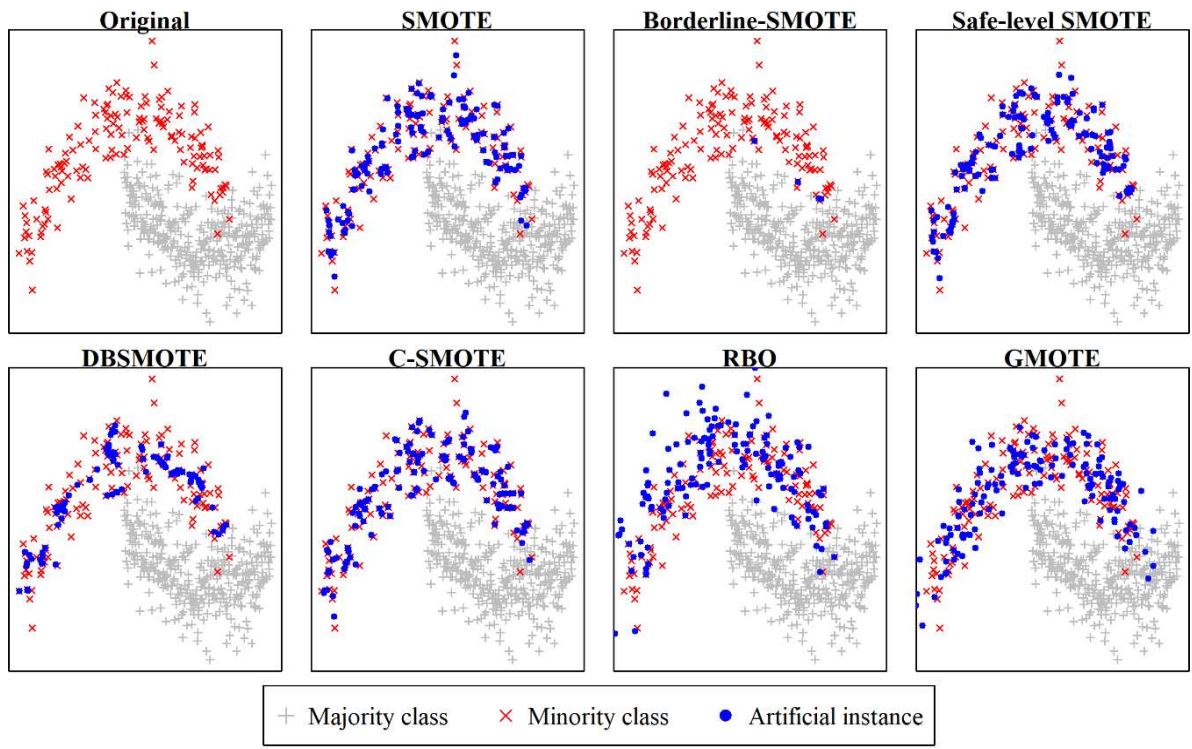

Fig. 4. An example 2 of artificial instances generation by different methods. The minority class and the majority class overlap slightly and have concave distributions.

To compare the model performance of GMOTE to other competing methods, we made a pair of GMOTE and each competing method. For the accuracy and F1-score of each pair, we conducted two types of paired Wilcoxon tests. The first test is to see if the score or rank of GMOTE is higher than the score or rank of the other method and the second one is to see if it is lower than the score or rank of the other method. The results of the tests are shown in Figs. 5 and 6 using the plus and minus signs above the boxplots. The plus sign indicates the score or rank of GMOTE is higher than the other while the minus sign indicates the opposite. The number of signs shows statistical significance where one, two, and three indicate that the p-value is less than 0.05, 0.01, and 0.001, respectively. And black diamond indicates the sample mean of two metrics for each method. The average metrics according to each method, dataset, and classifier are presented as tables in Appendix 1-6, and the best and worst results are highlighted in bold and italic, respectively. In Appendix 2, BLSMOTE and DBSMOTE combined with SVM achieve NA on KEEL2. NA means that all minority class is classified as the majority class.

As can be seen, GMOTE achieves robust performance in combination with CART and SVM. Combined with

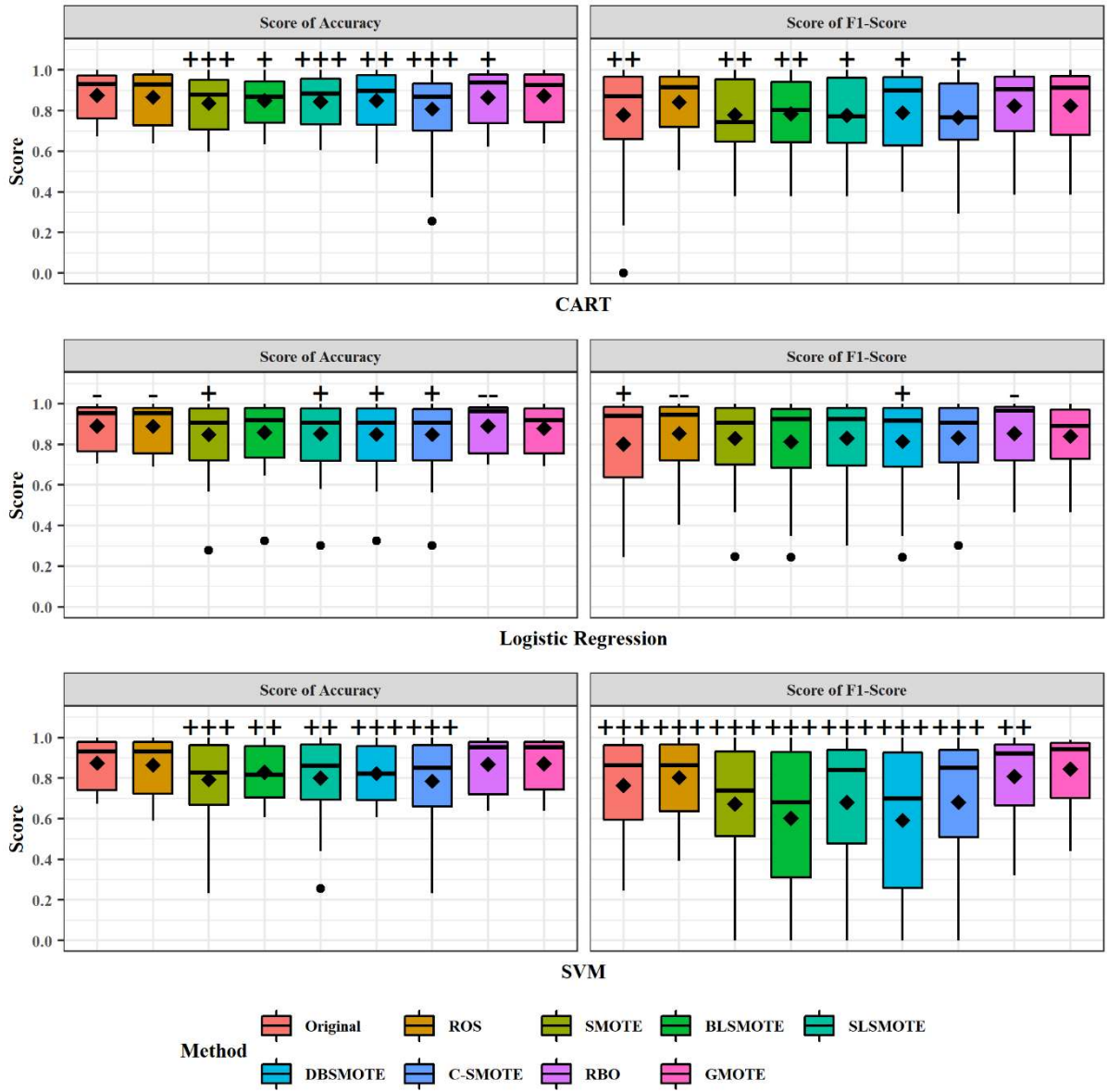

Fig. 5. The boxplots of score for accuracy and F1-Score.

CART or SVM, GMOTE achieves lower performance ranks on the precision. Since GMOTE would expand the training sample area of the minority class in the generating artificial instances process, it could increase the false-positive. Of course, conversely GMOTE has higher performance ranks on the recall. For example, in KEEL4, GMOTE combined with SVM achieves the lowest precision 0.849, but highest recall 0.943 (Appendix 3-4). Especially, GMOTE combined with SVM achieves a statistically higher score and rank of F1-Score than other oversampling methods (Figs. 5 and 6). On the other hand, the result shows that GMOTE performs slightly worse than RBO in logistic regression. Since RBO generates artificial instances in regions with a few majority class instances, it can thicken the interior of the kernel boundary of the minority class, unlike BLSMOTE, while making it as dense as other SMOTE methods. Therefore, rather than GMOTE, RBO may be appropriate for linear

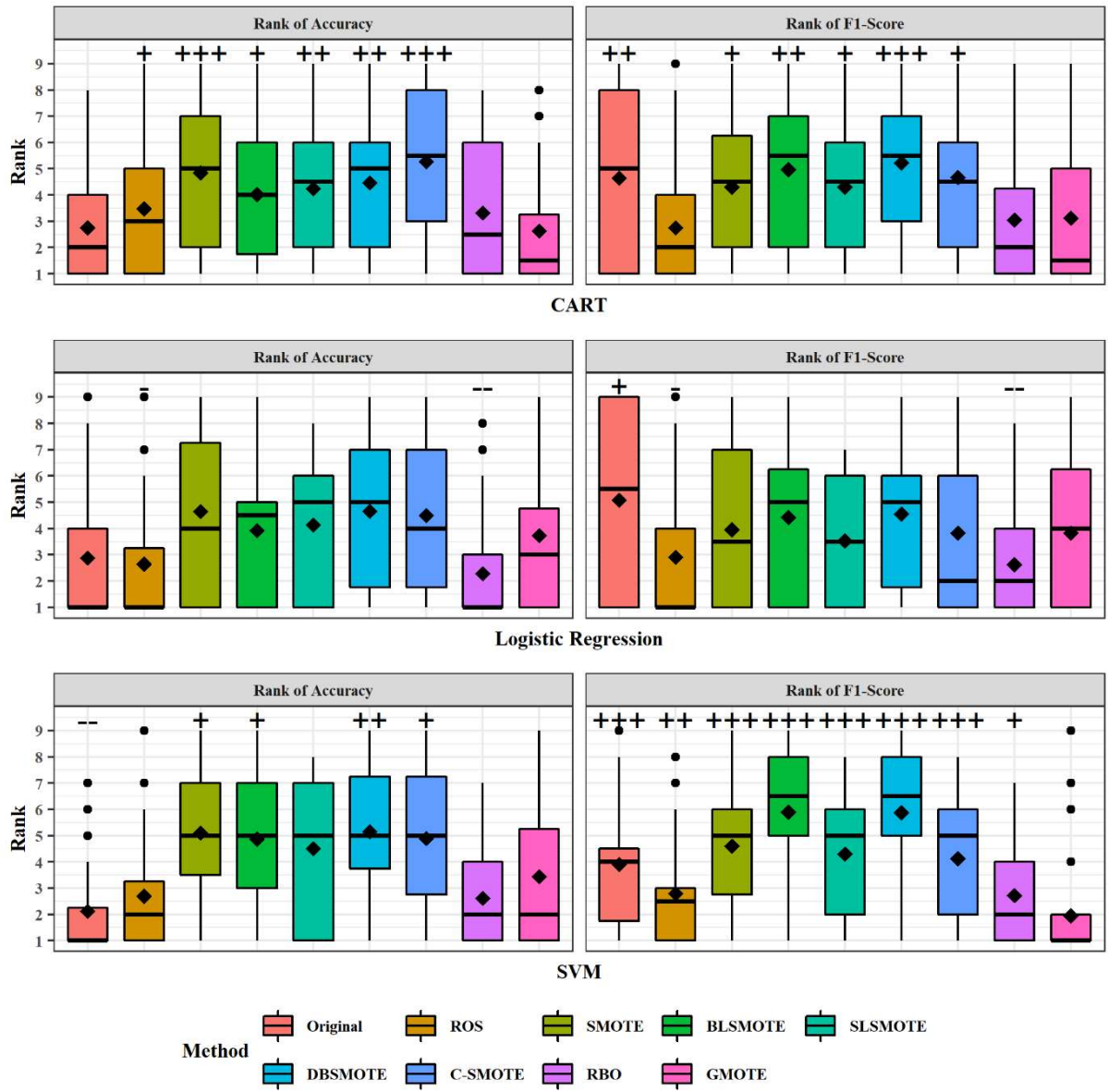

Fig. 6. The boxplots of rank for accuracy and F1-Score.

discriminant analysis such as logistic regression.

Overall, all methods score low on KEEL3 and KEEL8. For KEEL2, BLSMOTE and DBSMOTE do not predict the minority class at all. But GMOTE may be considered to achieve high rank and perform better than other methods to the rest of the dataset except KEEL4. These results show that GMOTE has robustness by reducing interference of local outliers in training sample space. Besides, GMOTE is appropriate when combined with classifiers handling collinearity or clusters in the distribution of minority classes.

## 5. Conclusions

In this paper, we propose the GMOTE as a new oversampling method with a statistical perspective for imbalanced datasets. We discussed the drawbacks of existing oversampling methods based on SMOTE approach. They were identified as depending on neighborhood analysis which does not take into account the distribution of minority class and local outlier. The proposed method utilizes the distribution-based generators by considering local outliers. First, we construct the GMM to estimate the distribution of minority class. In this process, the Mahalanobis distance is used to detect the local outlier from each mixture of GMM. Then, we generate artificial instances from the new GMM where local outliers were removed. These works reduce the influence of outliers and the bias of trained classifiers. To test the performance of GMOTE, we use the datasets from KEEL and toy datasets. Comprehensive experiments prove that the proposed method provides more natural instances generation and robust performance of Decision tree and SVM.

**Acknowledgements**

This work was supported by Basic Science Research Program through the National Research Foundation of Korea (NRF) funded by the Ministry of Education (NRF-2020R1A6A1A06046728).

Appendix

Appendix. 1
Averages of Accuracy

| Classifiers | Methods | Dataset | | | | | | | |
|---|---|---|---|---|---|---|---|---|---|
| | | KEEL1 | KEEL2 | KEEL3 | KEEL4 | KEEL5 | KEEL6 | KEEL7 | KEEL8 |
| CART | Original | 0.986 | 0.916 | 0.696 | 0.940 | **0.763** | **0.990** | 0.947 | **0.765** |
| | ROS | 0.986 | 0.920 | 0.683 | **0.972** | 0.708 | 0.989 | 0.939 | 0.741 |
| | SMOTE | *0.968* | 0.879 | 0.666 | 0.926 | *0.689* | 0.931 | 0.950 | *0.681* |
| | BLSMOTE | *0.968* | 0.874 | **0.725** | *0.912* | 0.740 | *0.916* | 0.950 | 0.716 |
| | SLSMOTE | *0.968* | 0.902 | 0.670 | 0.926 | 0.707 | 0.931 | 0.946 | 0.702 |
| | DBSMOTE | *0.968* | 0.888 | 0.657 | 0.953 | 0.710 | 0.981 | 0.941 | 0.707 |
| | C-SMOTE | *0.968* | *0.646* | *0.653* | 0.916 | 0.724 | 0.929 | *0.937* | 0.702 |
| | RBO | 0.986 | 0.902 | 0.670 | 0.963 | 0.719 | 0.989 | **0.955** | 0.737 |
| | GMOTE | **0.991** | **0.930** | 0.722 | 0.940 | 0.720 | **0.990** | 0.950 | 0.748 |
| Logistic Regression | Original | **0.982** | 0.921 | 0.742 | **0.991** | **0.768** | 0.996 | *0.966* | **0.757** |
| | ROS | 0.977 | 0.935 | **0.755** | **0.991** | 0.740 | **0.998** | 0.971 | 0.751 |
| | SMOTE | 0.959 | 0.781 | *0.725* | 0.977 | 0.728 | *0.985* | **0.972** | *0.662* |
| | BLSMOTE | 0.964 | 0.800 | 0.732 | 0.981 | 0.741 | 0.989 | *0.966* | 0.703 |
| | SLSMOTE | 0.964 | 0.790 | 0.735 | 0.977 | 0.728 | 0.988 | 0.968 | 0.674 |
| | DBSMOTE | 0.964 | 0.795 | 0.729 | 0.977 | *0.710* | 0.987 | *0.966* | 0.672 |
| | C-SMOTE | *0.955* | *0.776* | 0.732 | 0.981 | 0.730 | *0.985* | **0.972** | 0.668 |
| | RBO | **0.982** | **0.939** | **0.755** | **0.991** | 0.742 | 0.996 | **0.972** | 0.750 |
| | GMOTE | **0.982** | 0.911 | 0.751 | *0.935* | 0.740 | 0.997 | 0.969 | 0.747 |
| SVM | Original | **0.977** | 0.916 | **0.719** | 0.967 | 0.702 | 0.994 | 0.958 | **0.763** |
| | ROS | 0.968 | 0.911 | 0.670 | 0.972 | 0.703 | **0.995** | 0.959 | 0.741 |
| | SMOTE | *0.905* | 0.515 | *0.644* | **0.977** | 0.674 | 0.974 | **0.960** | *0.691* |
| | BLSMOTE | 0.918 | 0.748 | 0.702 | 0.972 | 0.655 | *0.972* | *0.956* | 0.714 |
| | SLSMOTE | 0.914 | 0.530 | 0.670 | 0.972 | 0.678 | 0.975 | **0.960** | 0.703 |
| | DBSMOTE | 0.914 | 0.743 | 0.673 | 0.967 | *0.646* | 0.973 | 0.958 | 0.711 |
| | C-SMOTE | 0.914 | *0.446* | 0.647 | **0.977** | 0.674 | 0.974 | **0.960** | 0.697 |
| | RBO | 0.973 | 0.925 | 0.686 | 0.963 | 0.707 | 0.994 | **0.960** | 0.744 |
| | GMOTE | **0.977** | **0.939** | 0.693 | *0.958* | **0.719** | 0.975 | **0.960** | 0.744 |

Appendix. 2
Averages of F1-Score. NA means that prediction of the minority class is failed.

| Classifiers | Methods | Dataset | | | | | | | |
|---|---|---|---|---|---|---|---|---|---|
| | | KEEL1 | KEEL2 | KEEL3 | KEEL4 | KEEL5 | KEEL6 | KEEL7 | KEEL8 |
| CART | Original | 0.979 | 0.808 | *0.267* | 0.779 | 0.637 | 0.965 | 0.924 | 0.501 |
| | ROS | 0.979 | 0.838 | 0.445 | **0.907** | 0.638 | 0.961 | 0.914 | **0.588** |
| | SMOTE | 0.958 | 0.721 | 0.444 | *0.628* | 0.604 | 0.784 | 0.931 | 0.516 |
| | BLSMOTE | *0.956* | 0.702 | 0.427 | 0.712 | 0.612 | *0.747* | 0.928 | 0.501 |
| | SLSMOTE | *0.956* | 0.766 | 0.394 | *0.628* | 0.599 | 0.786 | 0.924 | 0.538 |
| | DBSMOTE | *0.956* | 0.736 | 0.320 | 0.820 | *0.580* | 0.935 | 0.917 | 0.499 |
| | C-SMOTE | *0.956* | *0.553* | 0.447 | 0.672 | 0.630 | 0.779 | *0.911* | 0.520 |
| | RBO | 0.979 | 0.786 | 0.384 | 0.887 | **0.639** | 0.962 | **0.936** | 0.521 |
| | GMOTE | **0.986** | **0.871** | **0.448** | 0.795 | 0.630 | **0.967** | 0.930 | *0.482* |
| Logistic Regression | Original | **0.974** | 0.826 | *0.217* | **0.971** | 0.631 | 0.986 | *0.952* | *0.445* |
| | ROS | 0.968 | 0.862 | 0.423 | **0.971** | 0.664 | **0.992** | 0.959 | **0.594** |
| | SMOTE | 0.946 | 0.720 | 0.461 | 0.940 | 0.656 | 0.952 | **0.961** | 0.578 |
| | BLSMOTE | 0.953 | 0.741 | 0.352 | 0.949 | *0.615* | 0.964 | *0.952* | 0.533 |
| | SLSMOTE | 0.953 | 0.730 | 0.454 | 0.940 | 0.641 | 0.960 | 0.954 | 0.578 |
| | DBSMOTE | 0.953 | 0.729 | 0.350 | 0.940 | *0.615* | 0.959 | *0.952* | 0.575 |
| | C-SMOTE | *0.942* | *0.707* | **0.476** | 0.949 | 0.655 | *0.950* | **0.961** | 0.582 |
| | RBO | **0.974** | **0.873** | 0.434 | **0.971** | 0.666 | 0.985 | **0.961** | 0.573 |
| | GMOTE | **0.974** | 0.809 | 0.420 | *0.826* | 0.660 | 0.988 | 0.956 | 0.590 |
| SVM | Original | 0.966 | 0.784 | *0.189* | 0.892 | 0.456 | 0.977 | 0.942 | 0.476 |
| | ROS | 0.953 | 0.781 | 0.320 | 0.910 | 0.535 | **0.981** | 0.944 | 0.537 |
| | SMOTE | *0.881* | 0.419 | 0.354 | **0.929** | 0.409 | 0.899 | **0.946** | 0.240 |
| | BLSMOTE | 0.891 | *NA* | 0.236 | 0.912 | 0.235 | *0.892* | *0.941* | 0.224 |
| | SLSMOTE | 0.890 | 0.481 | 0.346 | 0.912 | 0.386 | 0.903 | **0.946** | 0.250 |
| | DBSMOTE | 0.882 | *NA* | 0.238 | 0.897 | *0.169* | 0.894 | 0.942 | *0.219* |
| | C-SMOTE | 0.892 | 0.500 | 0.364 | **0.929** | 0.413 | 0.899 | **0.946** | 0.245 |
| | RBO | 0.961 | 0.817 | 0.321 | *0.879* | 0.542 | 0.978 | **0.946** | 0.558 |
| | GMOTE | **0.968** | **0.883** | **0.387** | 0.885 | **0.659** | 0.918 | **0.946** | **0.562** |

Appendix. 3
Averages of Recall

| Classifiers | Methods | Dataset | | | | | | | |
|---|---|---|---|---|---|---|---|---|---|
| | | KEEL1 | KEEL2 | KEEL3 | KEEL4 | KEEL5 | KEEL6 | KEEL7 | KEEL8 |
| CART | Original | *0.960* | 0.782 | *0.173* | 0.714 | 0.593 | 0.964 | *0.916* | *0.410* |
| | ROS | *0.960* | 0.862 | 0.492 | 0.857 | **0.735** | **0.970** | 0.933 | **0.639** |
| | SMOTE | **0.973** | 0.722 | 0.504 | *0.543* | 0.691 | 0.861 | **0.954** | 0.608 |
| | BLSMOTE | *0.960* | *0.662* | 0.407 | 0.771 | 0.605 | *0.857* | 0.921 | 0.499 |
| | SLSMOTE | *0.960* | 0.742 | 0.406 | *0.543* | 0.628 | 0.861 | 0.941 | 0.599 |
| | DBSMOTE | *0.960* | 0.722 | 0.307 | 0.800 | *0.586* | 0.945 | 0.925 | 0.506 |
| | C-SMOTE | *0.960* | 0.682 | **0.529** | 0.686 | 0.684 | 0.858 | 0.921 | 0.578 |
| | RBO | *0.960* | 0.744 | 0.404 | **0.886** | 0.713 | **0.970** | **0.954** | 0.499 |
| | GMOTE | **0.973** | **0.960** | 0.443 | 0.743 | 0.693 | 0.967 | 0.946 | 0.415 |
| Logistic Regression | Original | 0.973 | 0.807 | *0.135* | **0.971** | *0.571* | 0.988 | *0.950* | *0.340* |
| | ROS | 0.973 | 0.864 | 0.344 | **0.971** | 0.739 | 0.994 | **0.971** | 0.629 |
| | SMOTE | **0.987** | 0.825 | 0.456 | **0.971** | 0.762 | 0.997 | **0.971** | **0.788** |
| | BLSMOTE | **0.987** | 0.845 | 0.295 | **0.971** | 0.635 | **1.000** | *0.950* | 0.606 |
| | SLSMOTE | **0.987** | 0.845 | 0.431 | **0.971** | 0.725 | 0.997 | 0.958 | 0.767 |
| | DBSMOTE | **0.987** | 0.825 | 0.295 | **0.971** | 0.710 | 0.997 | 0.954 | 0.760 |
| | C-SMOTE | **0.987** | 0.845 | **0.468** | **0.971** | 0.755 | 0.997 | **0.971** | 0.786 |
| | RBO | 0.973 | **0.885** | 0.357 | **0.971** | 0.739 | *0.985* | **0.971** | 0.583 |
| | GMOTE | 0.973 | *0.785* | 0.344 | *0.914* | 0.727 | 0.988 | 0.962 | 0.629 |
| SVM | Original | 0.947 | 0.664 | *0.123* | *0.829* | 0.362 | 0.958 | **0.992** | 0.373 |
| | ROS | 0.947 | 0.684 | 0.295 | 0.886 | 0.492 | 0.967 | *0.988* | 0.520 |
| | SMOTE | **0.974** | 0.520 | **0.371** | 0.914 | 0.336 | 0.821 | **0.992** | 0.193 |
| | BLSMOTE | 0.934 | *0.000* | 0.185 | 0.886 | 0.157 | *0.809* | **0.992** | *0.163* |
| | SLSMOTE | **0.974** | 0.640 | 0.334 | 0.886 | 0.299 | 0.827 | **0.992** | 0.193 |
| | DBSMOTE | *0.921* | *0.000* | 0.196 | 0.857 | *0.113* | 0.812 | **0.992** | *0.163* |
| | C-SMOTE | **0.974** | 0.800 | 0.384 | 0.914 | 0.344 | 0.821 | **0.992** | 0.198 |
| | RBO | 0.960 | 0.760 | 0.296 | *0.829* | 0.500 | 0.964 | **0.992** | 0.562 |
| | GMOTE | 0.973 | **0.942** | 0.369 | **0.943** | **0.776** | **0.979** | **0.992** | **0.569** |

Appendix. 4
Averages of Precision

| Classifiers | Methods | Dataset | | | | | | | |
|---|---|---|---|---|---|---|---|---|---|
| | | KEEL1 | KEEL2 | KEEL3 | KEEL4 | KEEL5 | KEEL6 | KEEL7 | KEEL8 |
| CART | Original | **1.000** | **0.876** | *0.316* | 0.927 | **0.691** | 0.967 | 0.933 | **0.650** |
| | ROS | **1.000** | 0.835 | 0.415 | 0.971 | 0.565 | 0.952 | *0.896* | 0.546 |
| | SMOTE | *0.950* | 0.783 | 0.403 | **1.000** | *0.553* | 0.727 | 0.909 | *0.474* |
| | BLSMOTE | 0.956 | 0.796 | **0.495** | *0.829* | 0.640 | 0.671 | **0.937** | 0.531 |
| | SLSMOTE | 0.956 | 0.844 | 0.392 | **1.000** | 0.619 | 0.730 | 0.908 | 0.508 |
| | DBSMOTE | 0.956 | 0.821 | 0.346 | 0.921 | 0.614 | 0.929 | 0.911 | 0.520 |
| | C-SMOTE | 0.956 | *0.521* | 0.389 | 0.900 | 0.613 | *0.720* | 0.902 | 0.500 |
| | RBO | **1.000** | 0.862 | 0.371 | 0.898 | 0.581 | 0.955 | 0.920 | 0.551 |
| | GMOTE | **1.000** | 0.815 | 0.458 | 0.865 | 0.586 | **0.967** | 0.916 | 0.602 |
| Logistic Regression | Original | **0.978** | 0.861 | **0.587** | 0.975 | **0.710** | 0.985 | **0.954** | **0.657** |
| | ROS | 0.965 | 0.870 | 0.557 | 0.975 | 0.605 | **0.991** | *0.947* | 0.563 |
| | SMOTE | 0.914 | 0.752 | 0.482 | 0.927 | 0.623 | 0.912 | 0.951 | *0.475* |
| | BLSMOTE | 0.928 | 0.762 | 0.482 | 0.940 | 0.658 | 0.931 | **0.954** | 0.511 |
| | SLSMOTE | 0.928 | 0.741 | 0.497 | 0.927 | 0.623 | 0.928 | 0.950 | 0.485 |
| | DBSMOTE | 0.928 | 0.762 | *0.477* | 0.927 | *0.592* | 0.925 | 0.950 | 0.483 |
| | C-SMOTE | *0.907* | *0.700* | 0.502 | 0.940 | 0.625 | *0.910* | 0.951 | 0.481 |
| | RBO | **0.978** | **0.871** | 0.559 | 0.975 | 0.611 | 0.985 | 0.951 | 0.567 |
| | GMOTE | **0.978** | 0.839 | 0.548 | *0.772* | 0.608 | 0.988 | 0.951 | 0.556 |
| SVM | Original | **0.988** | **0.980** | 0.457 | 0.967 | 0.625 | 0.997 | 0.898 | **0.661** |
| | ROS | 0.963 | 0.922 | 0.353 | 0.942 | 0.589 | **0.997** | 0.905 | 0.556 |
| | SMOTE | *0.809* | 0.448 | 0.341 | 0.950 | 0.559 | 0.996 | 0.905 | *0.433* |
| | BLSMOTE | 0.864 | *0.000* | 0.368 | 0.943 | 0.547 | 0.996 | *0.895* | 0.476 |
| | SLSMOTE | 0.824 | 0.432 | 0.363 | 0.943 | 0.580 | 0.996 | 0.905 | 0.482 |
| | DBSMOTE | 0.864 | *0.000* | *0.311* | 0.943 | *0.461* | 0.996 | 0.898 | 0.456 |
| | C-SMOTE | 0.827 | 0.284 | 0.347 | 0.950 | 0.563 | 0.996 | 0.905 | 0.439 |
| | RBO | 0.966 | 0.905 | 0.367 | 0.938 | 0.596 | 0.994 | 0.905 | 0.557 |
| | GMOTE | 0.964 | 0.845 | 0.411 | *0.849* | 0.573 | *0.865* | 0.905 | 0.556 |

Appendix. 5
Averages of G-Mean

| Classifiers | Methods | Dataset | | | | | | | |
|---|---|---|---|---|---|---|---|---|---|
| | | KEEL1 | KEEL2 | KEEL3 | KEEL4 | KEEL5 | KEEL6 | KEEL7 | KEEL8 |
| CART | Original | 0.980 | 0.857 | *0.333* | 0.828 | 0.711 | 0.979 | 0.940 | 0.609 |
| | ROS | 0.980 | 0.896 | 0.600 | 0.922 | 0.714 | **0.981** | 0.937 | **0.706** |
| | SMOTE | 0.968 | 0.805 | 0.599 | *0.689* | 0.681 | 0.898 | 0.951 | 0.639 |
| | BLSMOTE | *0.966* | 0.780 | 0.567 | 0.813 | 0.694 | *0.888* | 0.943 | 0.625 |
| | SLSMOTE | *0.966* | 0.830 | 0.552 | *0.689* | 0.674 | 0.898 | 0.945 | 0.658 |
| | DBSMOTE | *0.966* | 0.810 | 0.483 | 0.869 | *0.663* | 0.966 | 0.937 | 0.621 |
| | C-SMOTE | *0.966* | *0.622* | **0.605** | 0.762 | 0.703 | 0.895 | *0.933* | 0.643 |
| | RBO | 0.980 | 0.839 | 0.545 | **0.930** | **0.716** | **0.981** | **0.954** | 0.643 |
| | GMOTE | **0.986** | **0.938** | 0.591 | 0.848 | 0.708 | 0.980 | 0.949 | *0.600* |
| Logistic Regression | Original | **0.979** | 0.876 | *0.355* | **0.982** | 0.705 | 0.993 | *0.962* | *0.559* |
| | ROS | 0.976 | 0.908 | 0.551 | **0.982** | 0.738 | **0.996** | 0.971 | **0.709** |
| | SMOTE | 0.965 | *0.732* | 0.604 | 0.974 | 0.717 | 0.990 | **0.972** | 0.683 |
| | BLSMOTE | 0.968 | 0.766 | 0.490 | 0.977 | 0.694 | 0.994 | *0.962* | 0.657 |
| | SLSMOTE | 0.968 | 0.754 | 0.592 | 0.974 | 0.709 | 0.992 | 0.966 | 0.686 |
| | DBSMOTE | 0.968 | 0.756 | 0.488 | 0.974 | *0.687* | 0.991 | 0.963 | 0.683 |
| | C-SMOTE | *0.961* | 0.745 | **0.616** | 0.977 | 0.718 | *0.990* | **0.972** | 0.687 |
| | RBO | **0.979** | **0.918** | 0.562 | **0.982** | **0.739** | 0.991 | **0.972** | 0.689 |
| | GMOTE | **0.979** | 0.863 | 0.551 | *0.924* | 0.735 | 0.993 | 0.968 | 0.706 |
| SVM | Original | 0.969 | 0.809 | *0.335* | 0.907 | 0.563 | 0.978 | 0.965 | 0.585 |
| | ROS | 0.962 | 0.817 | 0.482 | 0.934 | 0.632 | **0.983** | 0.965 | 0.657 |
| | SMOTE | 0.918 | 0.217 | 0.522 | 0.950 | 0.526 | 0.905 | **0.967** | 0.379 |
| | BLSMOTE | 0.919 | *0.000* | 0.383 | 0.935 | 0.372 | *0.899* | *0.964* | 0.348 |
| | SLSMOTE | 0.926 | 0.308 | 0.508 | 0.935 | 0.505 | 0.909 | **0.967** | 0.389 |
| | DBSMOTE | *0.911* | *0.000* | 0.394 | 0.920 | *0.303* | 0.900 | 0.965 | *0.341* |
| | C-SMOTE | 0.925 | 0.262 | 0.532 | 0.950 | 0.529 | 0.905 | **0.967** | 0.383 |
| | RBO | 0.969 | 0.855 | 0.480 | *0.905* | 0.638 | 0.981 | **0.967** | 0.676 |
| | GMOTE | **0.976** | **0.938** | **0.543** | **0.951** | **0.730** | 0.976 | **0.967** | **0.680** |

Appendix. 6
Averages of AUC

| Classifiers | Methods | Dataset | | | | | | | |
|---|---|---|---|---|---|---|---|---|---|
| | | KEEL1 | KEEL2 | KEEL3 | KEEL4 | KEEL5 | KEEL6 | KEEL7 | KEEL8 |
| CART | Original | 0.980 | 0.929 | 0.626 | 0.887 | **0.793** | 0.983 | 0.953 | 0.724 |
| | ROS | 0.980 | **0.939** | 0.656 | 0.913 | 0.762 | 0.984 | 0.944 | **0.740** |
| | SMOTE | 0.969 | *0.782* | 0.612 | 0.900 | 0.737 | 0.889 | **0.960** | *0.667* |
| | BLSMOTE | *0.966* | 0.881 | 0.681 | **0.944** | 0.775 | *0.885* | 0.945 | 0.705 |
| | SLSMOTE | *0.966* | 0.892 | 0.636 | *0.839* | 0.747 | 0.889 | 0.946 | 0.693 |
| | DBSMOTE | *0.966* | 0.883 | 0.620 | 0.849 | *0.727* | 0.963 | *0.941* | 0.691 |
| | C-SMOTE | *0.966* | 0.800 | *0.604* | 0.915 | 0.747 | 0.886 | 0.944 | 0.675 |
| | RBO | 0.980 | 0.931 | 0.616 | 0.932 | 0.765 | 0.985 | 0.957 | 0.694 |
| | GMOTE | **0.987** | 0.937 | **0.686** | 0.900 | 0.766 | **0.986** | 0.948 | 0.728 |
| Logistic Regression | Original | **0.992** | 0.968 | 0.670 | **1.000** | **0.832** | *0.998* | **0.995** | **0.790** |
| | ROS | **0.992** | 0.969 | 0.693 | **1.000** | **0.832** | 0.999 | **0.995** | 0.788 |
| | SMOTE | 0.990 | 0.960 | 0.689 | *0.979* | 0.823 | 0.999 | *0.994* | 0.788 |
| | BLSMOTE | 0.990 | 0.953 | 0.687 | *0.979* | 0.822 | 0.999 | **0.995** | 0.789 |
| | SLSMOTE | 0.990 | 0.969 | 0.687 | *0.979* | 0.822 | 0.999 | *0.994* | 0.789 |
| | DBSMOTE | 0.990 | 0.964 | 0.686 | *0.979* | 0.816 | 0.999 | **0.995** | *0.786* |
| | C-SMOTE | *0.989* | 0.961 | 0.686 | 0.980 | 0.822 | 0.999 | *0.994* | 0.788 |
| | RBO | **0.992** | 0.963 | 0.672 | **1.000** | 0.829 | *0.998* | *0.994* | 0.788 |
| | GMOTE | **0.992** | *0.921* | *0.657* | 0.990 | 0.827 | **1.000** | *0.994* | **0.790** |
| SVM | Original | 0.988 | 0.985 | 0.640 | 0.996 | 0.770 | **0.999** | *0.986* | 0.744 |
| | ROS | 0.988 | 0.985 | 0.645 | **0.998** | 0.757 | **0.999** | 0.987 | 0.747 |
| | SMOTE | 0.970 | 0.804 | 0.650 | 0.991 | *0.703* | **0.999** | **0.988** | *0.651* |
| | BLSMOTE | 0.971 | 0.796 | 0.663 | 0.991 | 0.713 | **0.999** | *0.986* | 0.658 |
| | SLSMOTE | 0.970 | *0.779* | 0.654 | *0.990* | 0.715 | **0.999** | *0.986* | 0.660 |
| | DBSMOTE | 0.969 | 0.781 | *0.630* | *0.990* | 0.705 | **0.999** | 0.987 | 0.658 |
| | C-SMOTE | *0.967* | *0.779* | 0.643 | 0.991 | 0.709 | **0.999** | **0.988** | 0.660 |
| | RBO | 0.988 | **0.988** | **0.670** | 0.996 | 0.761 | **0.999** | 0.987 | **0.751** |
| | GMOTE | **0.989** | 0.973 | 0.663 | 0.997 | **0.787** | *0.997* | *0.986* | 0.749 |